\pdfoutput=1

\documentclass[11pt]{article}

\usepackage{acl}

\usepackage{times}
\usepackage{latexsym}

\usepackage[T1]{fontenc}

\usepackage[utf8]{inputenc}

\usepackage{microtype}
\usepackage{inconsolata}

\usepackage{subcaption} 
\usepackage{graphicx}
\usepackage[most]{tcolorbox}
\usepackage{verbatim}
\usepackage{comment}
\usepackage{booktabs}
\usepackage{float}  
\usepackage{adjustbox}

\usepackage{listings}

\usepackage{multirow}

\usepackage{xcolor}
\definecolor{pastelFrame}{RGB}{173,216,230} 
\definecolor{pastelTitle}{RGB}{213,234,249} 
\definecolor{pastelBack}{RGB}{250,250,250}  

\lstdefinestyle{optuna}{
  backgroundcolor=\color{white},
  basicstyle=\ttfamily\footnotesize,
  breaklines=true,
  keywordstyle=\color{mygreen}\bfseries,
  commentstyle=\color{mygray},
  stringstyle=\color{mymauve},
  showstringspaces=false
}

\title{o-MEGA: Optimized Methods for Explanation Generation and Analysis}

\author{Ľuboš Kriš$^{\spadesuit}$ \qquad Jaroslav Kopčan$^{\spadesuit}$ \qquad Qiwei Peng$^{\heartsuit}$ \\ \textbf{Andrej Ridzik}$^{\spadesuit}$ \qquad \textbf{Marcel Veselý}$^{\spadesuit}$ \qquad \textbf{Martin Tamajka}$^{\spadesuit}$   \\
$\spadesuit$Kempelen Institute of Intelligent Technologies \\ $\heartsuit$University of Copenhagen \\
\{name.surname\}@kinit.sk$^{\spadesuit}$ \qquad qipe@di.ku.dk$^{\heartsuit}$
}

\begin{document}
\maketitle
\begin{abstract}

The proliferation of transformer-based language models has revolutionized NLP domain while simultaneously introduced significant challenges regarding model transparency and trustworthiness. The complexity of achieving explainable systems in this domain is evidenced by the extensive array of explanation methods and evaluation metrics developed by researchers. To address the challenge of selecting optimal explainability approaches, we present \textbf{\texttt{o-mega}}, a hyperparameter optimization tool designed to automatically identify the most effective explainable AI methods and their configurations within the semantic matching domain. We evaluate o-mega on a post-claim matching pipeline using a curated dataset of social media posts paired with refuting claims. Our tool systematically explores different explainable methods and their hyperparameters, demonstrating improved transparency in automated fact-checking systems. As a result, such automated optimization of explanation methods can significantly enhance the interpretability of claim-matching models in critical applications such as misinformation detection, contributing to more trustworthy and transparent AI systems.


\end{abstract}

\section{Introduction}

The incredible success of today's transformer-based language models drastically changed the landscape of the natural language processing domain and how we approach complex tasks within it. However, with great power comes great responsibility. This success comes with a significant trade-off: the increasing complexity of these models has made them essentially black boxes, limiting their adoption in critical applications where transparency and interpretability are of great importance. Traditionally, addressing such interpretability challenges has relied on post-hoc relevance attribution methods, which provide post-hoc explanations \citep{lapuschkin2019unmasking,sogaard2021explainable} for trained models by assigning importance scores to input features or model parameters. While these explainable AI (XAI) techniques have proven valuable in revealing model behavior and identifying potential underlying flaws, so-called "Clever Hans" effects, or biases, they introduce a new challenge: the overwhelming variety of available XAI algorithms and their possible configurations. Just as practitioners may struggle to select optimal model architectures or hyperparameters, there is now emerging equally complex task of choosing the right explainability method for a given use case \citep{ali2023explainable}. This explainability challenge is especially relevant, for instance, in semantic matching tasks, where understanding why a model determined that two pieces of text are semantically similar or different is crucial for building user trust and ensuring reliable performance.
Moreover, the interpretable semantic matching scenario is relevant in situations where the nuanced relationships between two texts require explanations that describe the model's decision-making process precisely and understandably to end users. Different XAI algorithms may highlight different aspects of the input text, leading to varying levels of usefulness depending on the specific matching task, domain, and user requirements. The manual process of evaluating and selecting appropriate XAI configurations is time-consuming, resource-intensive, subjective, and often suboptimal.
Our \textbf{\texttt{o-mega}}\footnote{We have released the code as well as documentation and examples at \href{https://github.com/kinit-sk/o-mega}{o-mega repository}.} tool addresses this critical gap by automating the selection and optimization of post-hoc explainability algorithms, specifically within the semantic matching tasks. We have built the \textbf{\texttt{o-mega}} tool on top of previous work, which was inspired by the AutoML concept \citep{thornton2013auto}, and adjusted to the domain of explainable AI. The main functionality is focused on the systematic evaluation of different XAI methods and their configurations in order to identify those that provide the most useful explanations for a given model and dataset combination according to specified criteria.
The main motivation for the \textbf{\texttt{o-mega}} tool stems from the fact that explainability is not one-size-fits-all. In the semantic matching domain-whether it is a claim-matching task for fact-checking, document similarity assessment, or content recommendation-users need explanations that help them understand and trust the model's matching decisions. Since explainability is often overlooked because of its often ambiguous explanations, which are difficult to understand and struggles with implementation of the methods itself, by automating the explainability process of finding satisfactory XAI configurations, \textbf{\texttt{o-mega}} enables the use of interpretable semantic matching systems more efficiently while ensuring that the explanations provided are both technically correct and somehow practically useful. This tool represents an addition to the claim-matching task, which is central to fact-checking applications \citep{vo-lee-2020-facts,kazemi-etal-2021-claim,peng2025semeval}, and it is the first action step towards our effort to make it possible for domain experts to obtain meaningful insights from complex deep learning models without requiring extensive expertise in XAI methodologies.

\section{Related Work}
A wide range of explainable Artificial Intelligence (XAI) methods have been developed to enhance understanding of the decision-making process in machine learning models. Among post-hoc techniques, perturbation-based approaches are widely used where model predictions are examined by systematically sampling perturbed versions of the input and observing corresponding changes in output. LIME \citep{ribeiro2016should} provides local explanations by fitting an interpretable surrogate model to approximate the behavior of a complex model. SHAP \citep{lundberg2017unified} attributes feature importance based on Shapley values from cooperative game theory. Other perturbation-based methods include Occlusion Sensitivity \citep{zeiler2014visualizing}, which assesses feature relevance by masking parts of the input directly. In contrast, gradient-based methods such as Integrated Gradients \citep{sundararajan2017axiomatic} and LRP \citep{bach2015pixel} propagate relevance scores or gradients backward through the model to identify influential features. Additionally, some studies propose textual explanation generation methods that aim to produce human-readable justifications for model predictions \citep{lei-etal-2016-rationalizing,camburu2018snli,atanasova-etal-2020-generating-fact}. Explainability methods provide explanations of different qualities, and various metrics have been proposed to evaluate them. Common evaluation criteria include faithfulness, plausibility, and stability \citep{alvarez2018robustness,mohseni2021multidisciplinary,nauta2023anecdotal}. However, no single metric captures all aspects of explanation quality, and different metrics may yield conflicts. This makes the evaluation of XAI methods a persistent challenge. This challenge is further compounded when selecting the most appropriate explanation technique for a given task. To address this, AutoXAI frameworks \citep{autoxai2022}, inspired by AutoML systems \citep{he2021automl}, have been proposed to automate the selection and configuration of XAI techniques. The framework aims to adaptively choose the most suitable explanation method and optimize associated hyperparameters based on user-defined objectives. The existing AutoXAI framework works only with the tabular modality of data and has implemented only two methods - SHAP and LIME, which are often considered as baselines when dealing with structured data. Inspired by it, we have created the \textbf{\texttt{o-mega}} framework within the domain of information retrieval, focusing on textual data. oMEGA framework incorporates multiple XAI methods, metrics for evaluation, and puts a specific focus on the claim matching task, which is a crucial task in fact-checking.
 
\section{System Description}
\begin{figure*}[t]
  \centering
  \includegraphics[width=0.75\textwidth]{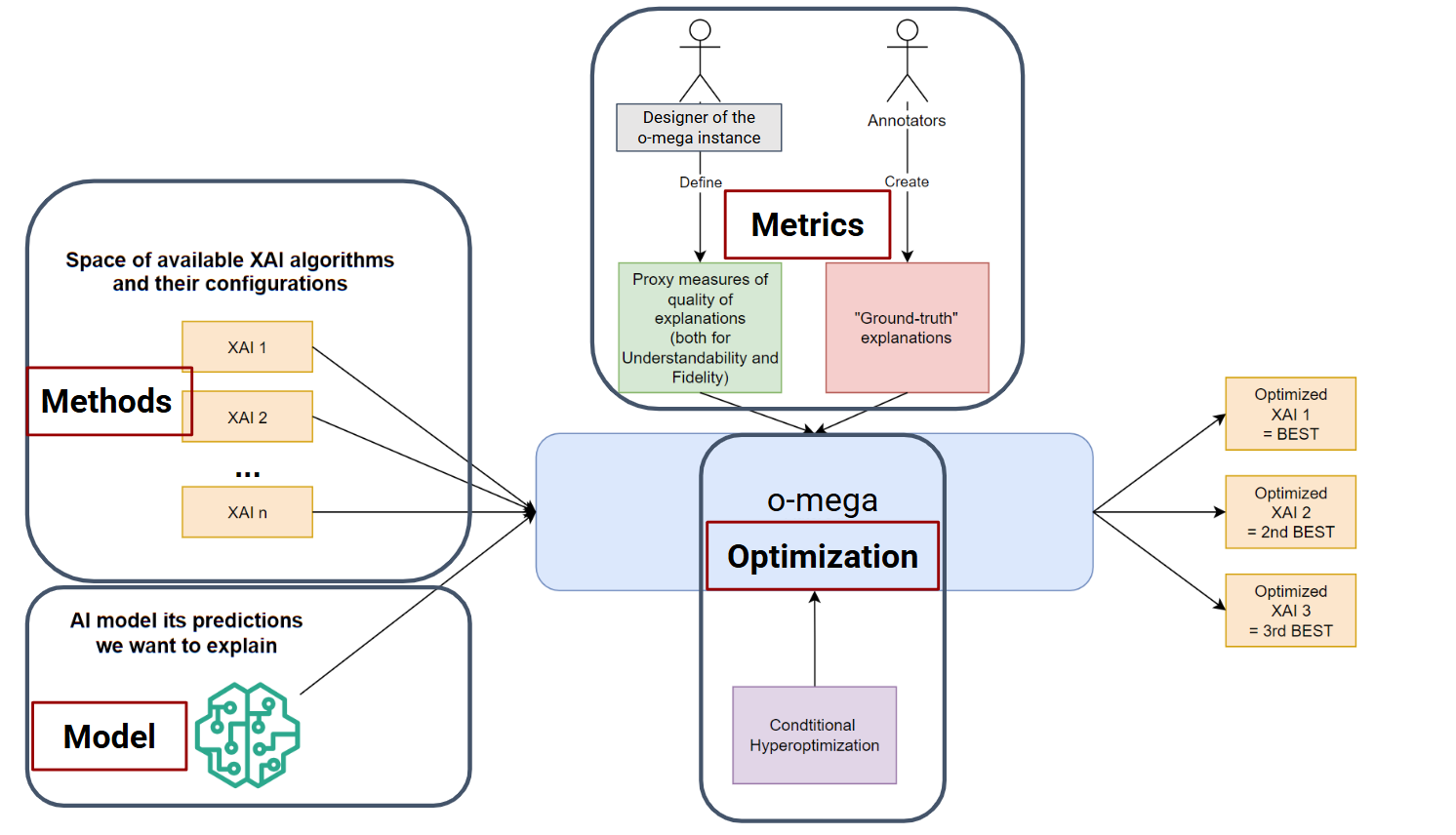}
  \caption{The architecture of the \textbf{\texttt{o-mega}} tool.}
  \label{fig:o-mega schema}
  \vspace{-10pt}
\end{figure*}

The \textbf{\texttt{o-mega}} tool is designed as a comprehensive framework for automatically selecting and optimizing explainable AI methods within semantic matching application. As illustrated in Figure \ref{fig:o-mega schema}, the system architecture consists of four interconnected modules that work together to identify the most effective explanation approach for a given AI model and dataset: 1) the \textbf{model} component that generates predictions requiring explanation, 2) the \textbf{method} component, which is a comprehensive space of available XAI algorithms and their configurations, 3) the \textbf{metric} part is an evaluation module incorporating both proxy measures and ground-truth annotations, and 4) the \textbf{optimization} component is a conditional hyperoptimization engine that systematically explores and ranks different explainability approaches.

\subsection{Methods}
For semantic matching and text classification tasks, we require attribution-based XAI methods capable of capturing token-level or word-level importance. Additionally, for semantic matching specifically, these methods must capture cross-sequence interactions as well.
We also want the methods to be diverse in their design and computationally efficient, therefore we have selected XAI methods spanning three paradigms: 1) gradient-based; 2) perturbation-based; and 3) architecture-specific. Overall, 11 explainable methods were implemented,  9 of them were from the Captum library. The entire selection is presented in Table \ref{tab:methods}. We have excluded several attribution methods from the library like Deconvolution, DeepLift, DeepLiftShap, because of irrelevancy for our task, or computational expenses.  

\subsection{Models}
The inherent design of how these post-hoc explainability methods are computed introduces significant architectural rigidness. This introduces a constraint to same extent - the scoping of supported models. We have included an exhaustive list of most widely used semantic matching models. Table \ref{tab:supported_architectures} presents the list of models which we have thoroughly tested. However, in general, the methods available within the \textbf{\texttt{o-mega}} framework should work with any transformer-based model for semantic matching of natural language text. 


\begin{table}[ht]
\centering
\resizebox{1\columnwidth}{!}{%
\begin{tabular}{l l}
\hline
\textbf{Model name} & \textbf{Embedding layer}\\
\hline
 sentence-transformers/gtr-t5-large & encoder.embed\_tokens \\
 sentence-transformers/gtr-t5-xl & encoder.embed\_tokens \\
 sentence-transformers/sentence-t5-xl & encoder.embed\_tokens \\
 sentence-transformers/all-mpnet-base-v2 & embeddings.word\_embeddings \\
 sentence-transformers/multi-qa-mpnet-base-cos-v1 & embeddings.word\_embeddings \\
 sentence-transformers/all-MiniLM-L12-v2 & embeddings.word\_embeddings \\
BAAI/bge-large-en-v1.5 & embeddings.word\_embeddings \\
BAAI/bge-base-en-v1.5 & embeddings.word\_embeddings \\
BAAI/bge-small-en-v1.5 & embeddings.word\_embeddings \\
llmrails/ember-v1 & embeddings.word\_embeddings \\
thenlper/gte-large & embeddings.word\_embeddings \\
intfloat/e5-large-v2 & embeddings.word\_embeddings \\
BAAI/bge-large-en-v1.5 & embeddings.word\_embeddings \\
\hline
\end{tabular}%
}
\caption{Overview of tested language models for explainable use in semantic matching}

\label{tab:supported_architectures}
\end{table}

\begin{table}[ht]
\centering
\resizebox{0.4\columnwidth}{!}{%
\begin{tabular}{p{4cm} p{2cm}}
\hline
\textbf{Method} \\
\hline
\multicolumn{2}{l}{\textit{Captum-based methods}} \\
\hline
Occlusion token-level \\
Input X Gradient  \\
Guided Backprop \\
Feature Ablation \\
Kernel Shap \\
Gradient Shap \\
LIME \\
Saliency \\
\hline
\multicolumn{2}{l}{\textit{Custom implementations}} \\
\hline
GAE  \\
Conservative LRP \\
Occlusion word-level \\
\hline
\end{tabular}%
}
\caption{Overview of available XAI methods}
\label{tab:methods}
\vspace{-10pt}
\end{table}

\subsection{Metrics}
Evaluation module of the \textbf{\texttt{o-mega}} tool is the most crucial one, because based on the measurements of quality of computed explanations, the optimization process is guided, and at the output ends the recommendation for best methods regarding the task, data and model used. In order to evaluate how good the provided explanations actually are, it is important to firstly define what is meant by \textit{ the explanation quality - what constitutes to a good explanation?}
There are two main approaches for how to answer this quality question, the first one is quantitative measurements, the second one is qualitative. While the qualitative measurements are just as much important, because of their subjective nature they can not be expressed by metrics. Therefore, within the o-mega evaluation we have focused primarily on the quantitative approach. On this topic was a lot of work done \citep{zhou2021intrinsic,markus2021role} about how to find out if the explanation is of high quality - it needs to adhere to two components - \textit{fidelity} and \textit{ plausibility}. 
\begin{itemize}
    \item \textbf{Explanation high in fidelity} reflects the actual underlying behavior of the model, meaning the features deemed as important by explanation method for specific prediction, should be the same which influenced model the most. Metrics designed to test this are based on ablations or perturbations.
    \item  \textbf{Explanation high in plausibility} reflects how easy is to comprehend the explanation for human-being, while this being the closest measure possible to qualitative evaluation within the quantitative domain. In order to be able express this semi-subjective property these metrics needs the human annotations - what humans themselves have deemed to be comprehensible explanation given the data and the task.
\end{itemize}
While during the evaluation is important to score high in both of these measurements, they often conflict with each other. An optimal explanation requires to strike a balance between them, while prioritizing one typically comes at the expense of the other. Our \textbf{\texttt{o-mega}} framework has available  metrics for both categories - fidelity and plausibility which we have deemed as important and user could choose before the optimization if he wants to favor one over the other, or optimize towards both of them. All these metrics are custom implemented versions of established, previously published metrics which work with textual data \citep{deyoung2020eraser,hedstrom2023quantus,attanasio-etal-2023-ferret}.

\begin{table}[ht]
\centering
\resizebox{0.85\columnwidth}{!}{%
\begin{tabular}{l c}
\hline
\textbf{Metrics} & \textbf{Category} \\
\hline
AOPC Comprehensiveness & \multirow{2}{*}{Fidelity} \\
AOPC Sufficiency &  \\
\hline
AUPRC & \multirow{4}{*}{Plausibility} \\
Token-level F1 scores & \\
Token-level IoU & \\
Average Precision Score & \\
\hline
\end{tabular}%
}
\caption{Overview of available metrics}
\label{tab:metrics}
\end{table}

\subsection{Optimization}
For hyperparameter optimization, we employed the Optuna library \citep{optuna2024}, which offers significant advantages over traditional grid search approaches. While grid search exhaustively evaluates all possible parameter combinations, Optuna uses intelligent search strategies such as Tree-structured Parzen Estimator (TPE) and Bayesian optimization that learn from previous evaluations to guide future parameter selection \citep{feurer2019hyperparameter}. This adaptive approach allows the optimizer to focus computational resources on promising regions of the hyperparameter space, effectively avoiding areas that have already shown poor performance.
The efficiency gains become particularly pronounced in high-dimensional hyperparameter spaces, where grid search suffers from the curse of dimensionality. For instance, if we have 5 hyperparameters with 10 possible values each, grid search would require $10^5$ = 100,000 evaluations, while Optuna's intelligent sampling can often find near-optimal configurations with significantly fewer trials often by an order of magnitude or more. This efficiency is crucial in our context, where Captum's explanation methods have numerous hyperparameters that can substantially impact both computational cost and explanation quality.
Moreover, Optuna's pruning capabilities allow early termination of unpromising trials, further reducing computational overhead. This is particularly valuable when working with explanation methods that may have expensive evaluation metrics, as the optimizer can quickly identify and abandon parameter configurations that are unlikely to yield good results based on partial evaluations. The result is a more efficient exploration of the hyperparameter space that converges to high-quality explanations faster than traditional optimization approaches.

\subsection{Configuration}
As a base part of the \textbf{\texttt{o-mega}}, the user can select a model from the Huggingface library (See Table~\ref{tab:supported_architectures}) or a local pre-trained model. Plausibility and faithfulness have corresponding weights to adjust the priority of each group of metrics (See Figure \ref{fig:model-explain-config}). Then, the form of evaluation of explanations can take place in 3 options: either in token form, or by combining explanations and masks into sentences, or into words. One example of the base configuration is shown in Figure \ref{fig:base-config}.

\begin{figure}[ht]
\centering
\begin{adjustbox}{width=1\linewidth}
\begin{lstlisting}[style=optuna, language=]
model_path: "intfloat/multilingual-e5-large"
embeddings_module_name: "embeddings.word_embeddings"
methods: ["GAE_Explain","Occlusion"]
normalizations: ["without_normalize"]
explanation_maps_token: True
plausability_weight: 0.5
faithfulness_weight: 0.5
multiple_object: False
\end{lstlisting}%
\end{adjustbox}
\caption{Configuration block specifying the base parameters for hyperoptimization and evaluation of explanations}
\label{fig:base-config}
\vspace{-10pt}
\end{figure}

The user can further configure 6 search strategies from the Optuna library (See Figure ~\ref{fig:optunaconfig}). These 6 search strategies can be divided into 2 groups: (1)single-object and (2)multi-object hyperoptimization. Within the configuration, plausibility and plausibility values can be considered separately. Within single-object hyperoptimization, we have TPESampler, GPSampler, and BruteForceSampler available for use, and within multi-object hyperoptimization, we can use NSGAIIISampler, NSGAIISampler, and TPESampler.

\begin{figure}[ht]
\centering
\begin{adjustbox}{width=0.5\linewidth}
\begin{lstlisting}[style=optuna, language=]
Optuna_parameters:
  sampler: "TPESampler"
  n_trials: 14
  n_startup_trials: 4
  seed: 1000
\end{lstlisting}
\end{adjustbox}
\caption{Configuration block for Optuna Sampler and its parameters.}
\label{fig:optunaconfig}
\vspace{-8pt}
\end{figure}

XAI methods also accept a list of configurations for customization. One example is shown in Figure \ref{fig:hyper-config}, where the hyperparameters of XAI methods can be restrained to a specific search space.

\begin{figure}[ht]
\centering
\begin{adjustbox}{width=0.9\linewidth}
\begin{lstlisting}[style=optuna, language=]
method_param:
  Gradient Shap:
    parameters:
      stdevs: (0.1, 0.9, {'step': 0.1})
      n_samples: [10, 15]
  Lime:
    parameters:
      n_samples: [80, 90]
  token_groups_for_feature_mask: true
\end{lstlisting}
\end{adjustbox}
\caption{Configuration block specifying hyperparameters for explainable method.}
\label{fig:hyper-config}
\vspace{-10pt}
\end{figure}

\section{Case Study}

\begin{figure}[t]
  \centering
  \includegraphics[width=\columnwidth]{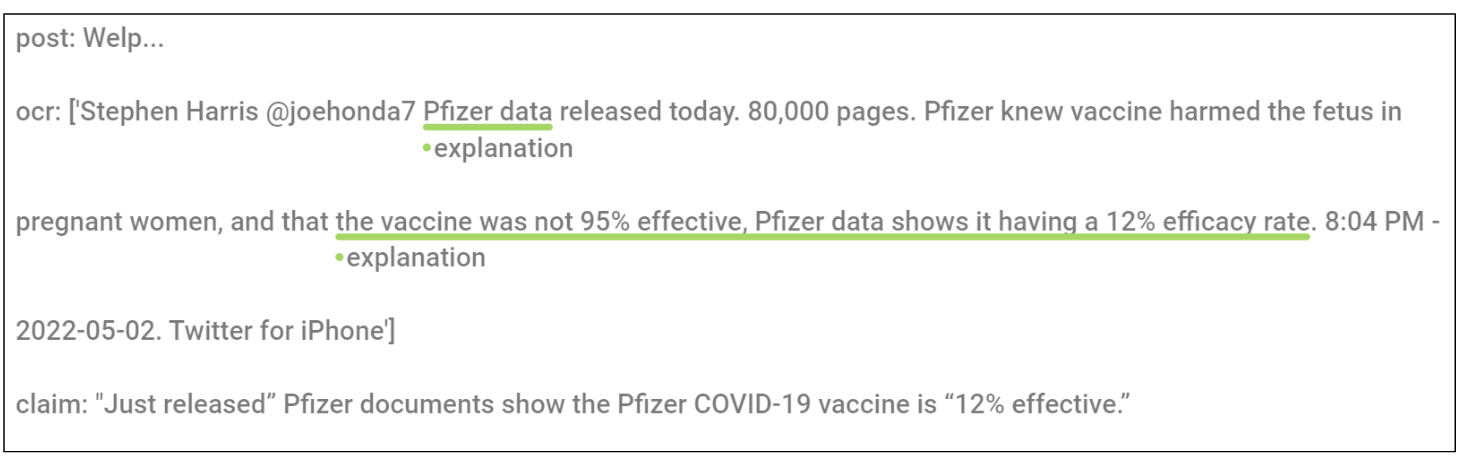}
  \caption{One example of the post-claim pair with annotated explanations.}
  \label{fig:doccano example}
  \vspace{-12pt}
\end{figure}

In this case study, we focus on the task of claim matching, a key component in automated fact checking. The claim matching task is typically formulated as an information retrieval task. Given an input text (e.g., social media posts), the goal is to find an appropriate claim from a database of claims that have already been fact-checked by professional fact-checkers. When a user posts a post making a claim worth fact-checking, the model aims to find a semantically similar claim from a list of previously fact-checked claims. For this study, we utilize the MultiClaim dataset \citep{pikuliak-etal-2023-multilingual}. To examine interpretability, five human annotators are asked to provide explanations for why specific claims were matched. Examples are given in figure \ref{fig:doccano example}. To assess the quality of the annotation, if the majority vote is not determined, we directly throw the sample away. This results in 512 post-claim pairs with human annotations.

\subsection{Task Configuration}
To generate optimal explanations for this task, we developed a hyperparameter optimization approach that selects the best combination of explanation methods and their configurations. The explanation methods identify significant attributes in both posts and claims by analyzing the cosine similarity between them. We utilized the Captum library for generating explanations, which provides multiple options for modifying explainable method parameters that affect how individual embeddings are processed. Our hyperparameter optimization framework incorporates these configurable parameters as well. Specific methods need to set the model in a specific way. In Figure~\ref{fig:model-explain-config} are shown the model parameters required for each explanation method are shown, such as the necessary layer or gradient settings, which show how different methods require access to specific components of the neural architecture. Figure~\ref{fig:method-explain-config} details the specific tunable parameters for each explanation method, such as perturbation sizes for occlusion methods or baseline values for gradient-based approaches. For the optimization process, we employed the Optuna library, which offers various optimization algorithms. Our evaluation framework uses five metrics which can be organized into two categories: fidelity and plausibility. Additionally, we included average precision score (APS) as a supplementary metric for plausibility assessment.

\subsection{Results and Analysis}
Table \ref{tab:comparison_methods} presents results from \textbf{\texttt{o-mega}} optimization which should provide clear guidance for practitioners. Occlusion is the recommended explainability method for this semantic matching task, achieving the best overall performance (0.819 average) by balancing technical accuracy with human interpretability. This automated recommendation eliminates the need for manual trial-and-error testing of different explanation methods. The results reveal that while most methods can accurately capture model behavior (faithfulness >0.94), they vary significantly in producing explanations that users can easily understand (plausibility 0.35-0.68). This means that method selection critically impacts user experience. Occlusion provides best explanations that are both technically correct and accessible to non-experts, while alternatives like ConservativeLRP (0.641) may confuse users despite being technically valid. The optimization process successfully identified the configuration that maximizes both explanation quality dimensions, providing practitioners with a data-driven recommendation rather than relying on popularity or default settings of explanation methods.

Table \ref{tab:sampler_comparison} depicts comparison of single objective samplers. Results reveals TPESampler as the most efficient optimization algorithm for this task, completing the same optimization quality in significantly less time compared to alternatives. While all three samplers identify Saliency as the optimal method with identical performance scores, TPESampler demonstrates superior computational efficiency, requiring only 14 trials versus BruteForceSampler's 35 trials to achieve the same result. BruteForceSampler provides the most thorough exploration but at a substantial computational cost (1.112 hours), making it impractical for real-world applications. GPSampler offers a middle ground with moderate efficiency (0.672 hours, 14 trials) but shows higher memory usage (760MB against 748MB for TPESampler) and experiences duplicate trials, indicating less efficient search space exploration. The main recommendation from the optimization run comparison is that the TPESampler represents the optimal choice, delivering the same explanation quality recommendations as exhaustive search methods while reducing optimization time by approximately 66\%, making automated XAI selection feasible for routine deployment. Lastly, Table \ref{tab:multi_objective_sampler_comparison} shows the multi-objective optimization results. It shows that BruteForceSampler and TPESampler both correctly identify Occlusion as the optimal method, while NSGA-II and NSGA-III variants select the inferior Saliency method. BruteForceSampler achieves the highest performance scores (0.957 faithfulness, 0.646 plausibility) but requires 35 trials and 1.121 hours, while TPESampler delivers nearly identical results (0.958 faithfulness, 0.622 plausibility) with 70\% less computation time (14 trials, 0.339 hours). The NSGA variants show identical performance, indicating no benefit from the increased complexity of NSGA-III. For practitioners, TPESampler is the recommended choice, providing the correct method selection with optimal efficiency for multi-objective explainability optimization.

\begin{table}[t]
\centering
\resizebox{0.8\columnwidth}{!}{%
\begin{tabular}{l| c c c}
\hline
\textbf{Methods} & \textbf{Faithfulness} & \textbf{Plausibility} & \textbf{Average} \\ \hline
Occlusion & 0.963 & \textbf{0.675} & \textbf{0.819} \\ 
Gradient Shap & \textbf{0.967} & 0.627 & 0.797 \\ 
Saliency & 0.966 & 0.621 & 0.794 \\ 
GAE\_Explain & 0.959 & 0.615 & 0.787 \\ 
Lime & 0.962 & 0.599 & 0.781 \\ 
Feature Ablation & 0.960 & 0.597 & 0.779 \\ 
Occlusion word-level & 0.946 & 0.605 & 0.776 \\ 
Kernel Shap & 0.952 & 0.563 & 0.757 \\ 
Guided Backprop & 0.943 & 0.516 & 0.730 \\ 
Input X Gradient & 0.935 & 0.497 & 0.716 \\ 
ConservativeLRP & 0.927 & 0.354 & 0.641 \\ 
\hline
\end{tabular}%
}
\caption{Comparison of Methods with Faithfulness, Plausibility, and Average scores.}
\label{tab:comparison_methods}
\vspace{-10pt}
\end{table}

\begin{table}[ht]
    \centering
    \resizebox{1\columnwidth}{!}{%
    \begin{tabular}{l c c c}
    \hline
    \textbf{Single objective} & \textbf{BruteForceSampler} & \textbf{GPSampler} & \textbf{TPESampler} \\
    \hline
    method\_best & Saliency & Saliency & Saliency \\
    overall\_score & 0.784 & 0.78 & 0.784 \\
    best\_find\_at & 13 & 5 & 7 \\
    peak memory usage (MB) & 936.53 & 760.92 & \textbf{748.43} \\
    time (hours) & 1.112 & 0.672 & \textbf{0.379} \\
    number\_dup & 0 & \textbf{1} & 4 \\
    all\_trials & 35 & 14 & 14 \\
    \hline
    \end{tabular}%
    }
    \caption{Comparison of single objective samplers' performance. The best performances are depicted in bold.}
    \label{tab:sampler_comparison}
    \vspace{-10pt}
\end{table}

\begin{table}[ht] 
\centering
\resizebox{1\columnwidth}{!}{%
\begin{tabular}{l c c c c }
\hline
\textbf{Metric} & \textbf{BruteForceSampler} & \textbf{TPESampler} & \textbf{NSGAIISampler} & \textbf{NSGAIIISampler} \\ \hline
Total Trials & 35 & \textbf{14} & \textbf{14} & \textbf{14} \\ 
Best Method & Occlusion & Occlusion & Saliency & Saliency \\
Faithfulness & 0.957 & \textbf{0.958} & 0.957 & 0.957 \\
Plausibility & \textbf{0.646} & 0.622 & 0.606 & 0.606 \\ 
Peak Memory (MB)& 907.08 & 761.84 & \textbf{747.27} & 749.35 \\ 
Time (hours) & 1.121 & 0.339 & \textbf{0.261} & \textbf{0.261} \\ 
Duplicates & 0 & 5 & 7 & 7 \\ 
\hline
\end{tabular}%
}
\caption{Comparison of Multi-Objective Samplers' Performance}
\label{tab:multi_objective_sampler_comparison}
\vspace{-10pt}
\end{table}


\section{Conclusion}
This work presents \textbf{\texttt{o-mega}}, an automated hyperparameter optimization tool for explainable AI method selection in semantic matching and text classification tasks. Our evaluation within semantic matching demonstrates that automated optimization successfully identifies optimal XAI configurations, with Occlusion emerging as the best-performing method for post-claim matching applications. By automating the complex process of XAI method selection and configuration, o-mega enables users to deploy transparent AI systems without requiring deep expertise in explainability techniques. This work has demonstrated the possibility of making explainability more accessible and frictionless for real-world applications, particularly in critical domains such as automated fact-checking and disinformation detection. 


\section*{Limitations}
The current implementation of o-mega has several limitations:
\begin{itemize}
    \item The tool was developed specifically for claim matching as an enhancer for a specific dataset - MultiClaim, although the text classification pipeline is also available, it is still limiting its immediate applicability to other domains 
    \item Evaluation metrics are currently restricted to only established ones, and tools do not include other experimental metric evaluations, such as localization and sparseness
    \item Computationally expensive explanation methods have not been incorporated yet, but this is also partially desired since high computational resource requirements render the methods less effective, therefore users are less likely to use them
\end{itemize}

\section*{Acknowledgments}
We would like to thank all reviewers for their insightful comments and feedback. This work was supported by DisAI - Improving scientific excellence and creativity in combating disinformation with artificial intelligence and language technologies, a project funded by European Union under the Horizon Europe, GA No. \href{https://doi.org/10.3030/101079164}{101079164}.
This work was also partially funded by European Union, under the project lorAI - Low Resource Artificial Intelligence, GA No. \href{https://doi.org/10.3030/101136646}{101136646}.

\bibliography{custom}

\appendix

\section*{Appendix: Method Hyperparameters}
\label{app:hyperparams}

\begin{table}[ht]
\centering
\renewcommand{\arraystretch}{1.2}
\small
{\scriptsize 
\begin{tabular}{l| p{4.5cm}}
\hline
\textbf{Methods} & \textbf{Hyperparameters} \\ \hline
Occlusion & sliding\_window\_shapes:(5, 1024), strides:(1, 1024) \\
Gradient Shap & stdevs:0.1, n\_samples:15 \\
Saliency & abs: True \\ 
GAE\_Explain & \\ 
Lime & n\_samples:90 ,similarity\_func:\{'function\_name': 'get\_exp\_kernel\_similarity\_function', 'parameters': \{'distance\_mode': 'euclidean', 'kernel\_width': 750\}\}, interpretable\_model:\{'function\_name': 'SkLearnLasso', 'parameters': \{'alpha': 1e-10\}\} \\ 
Feature Ablation & \\
Occlusion\_word\_level & regex\_condition: ".,!?;:…" \\ 
Kernel Shap & n\_samples:90 \\
Guided Backprop & \\ 
Input X Gradient & \\
ConservativeLRP & \\ 
\end{tabular}
}
\caption{Complete list of method hyperparameters}
\label{tab:app_hyperparams}
\end{table}

\lstset{
  breaklines=true,
  breakatwhitespace=false,
  columns=flexible
}

\begin{figure}[ht]
\centering
{\tiny
\begin{lstlisting}[style=optuna, language=]
model_param:
   Lime:
     similarity_func:
       function_name:
         - captum.attr._core.lime.get_exp_kernel_similarity_function
       parameters:
         distance_mode: ["cosine", "euclidean"]
         kernel_width: [450, 750]
     interpretable_model:
       function_name:
         - captum._utils.models.linear_model.SkLearnLasso
       parameters:
         alpha: [1e-19, 1e-25]
   GAE_Explain:
     implemented_method: true
     layers:
       module_path_expressions:
         - "hf_transformer.encoder.layer.*.attention.self.dropout"
   ConservativeLRP:
     implemented_method: true
     layers:
       store_A_path_expressions:
         - "hf_transformer.embeddings"
       attent_path_expressions:
         - "hf_transformer.encoder.layer.*.attention.self.dropout"
       norm_layer_path_expressions:
         - "hf_transformer.embeddings.LayerNorm"
         - "hf_transformer.encoder.layer.*.attention.output.LayerNorm"
         - "hf_transformer.encoder.layer.*.output.LayerNorm"
   Occlusion_word_level:
     implemented_method: true
\end{lstlisting}
}
\caption{Configuration block specifying the model parameters for hyperoptimization and evaluation of explanations}
\label{fig:model-explain-config}
\end{figure}

\begin{figure}[ht]
\centering
{\tiny
\begin{lstlisting}[style=optuna, language=]
 
 method_param:
   Lime:
     parameters:
       n_samples: [80, 90]
     token_groups_for_feature_mask: true
   Saliency:
     parameters:
       abs: [true, false]
   Occlusion:
     parameters:
       sliding_window_shapes:
         - [3, 1024]
         - [5, 1024]
       strides:
         - [1, 1024]
         - [1, 512]
     compute_baseline: true
   Gradient Shap:
     parameters:
       stdevs: [0.1, 0.9]
       n_samples: [10, 15]
     compute_baseline: true
   Kernel Shap:
     parameters:
       n_samples: [80, 90]
     compute_baseline: true
   Feature Ablation:
     token_groups_for_feature_mask: true
     compute_baseline: true
   Occlusion_word_level:
     parameters:
       regex_condition:
         - ""
         - ".,!?;:"
   Integrated Gradients:
     parameters:
       n_steps: [60, 40]
\end{lstlisting}
}
\caption{Configuration block specifying the method parameters for hyper-optimization and evaluation of explanations}
\label{fig:method-explain-config}
\end{figure}
\end{document}